%% file: CaRoSaC.tex
\documentclass[letterpaper, 10 pt, journal, twoside]{IEEEtran}

\IEEEoverridecommandlockouts 
\usepackage{array} 
\usepackage[utf8]{inputenc}
\usepackage[english]{babel}
\usepackage{xcolor} 
\usepackage{setspace}
\usepackage{cite}
\usepackage{multirow}
\usepackage{subfigure}
\usepackage{graphics} 
\usepackage{graphicx}
\graphicspath{ {./images/} }
\usepackage[mathscr]{euscript}
\usepackage{url}
\usepackage{tikz}
\usetikzlibrary{intersections}
\usetikzlibrary{decorations.pathreplacing}
\usepackage[infoshow,debugshow]{tabularx}
\usepackage{booktabs}
\usepackage{textcomp}
\usepackage[hidelinks]{hyperref}

\usepackage{multicol}
\usepackage[noend]{algpseudocode}

\usepackage{amsthm} 
\usepackage{amsmath} 
\usepackage{amssymb}  
\usepackage[percent]{overpic}
\usepackage[ruled,vlined]{algorithm2e}
\usepackage[noend]{algpseudocode}

\theoremstyle{plain}

\usepackage{enumitem}
\setitemize{noitemsep,topsep=0pt,parsep=0pt,partopsep=0pt}

\renewcommand{\baselinestretch}{0.95}

\title{\LARGE \bf{CaRoSaC: A Reinforcement Learning-Based
Kinematic Control of Cable-Driven Parallel Robots by Addressing Cable Sag through
Simulation}}
\author{Rohit Dhakate$^{1}$,  {Thomas Jantos}$^{1}$, {Eren Allak}$^{1}$, {Stephan Weiss}$^{1}$, and {Jan Steinbrener}$^{1}$
\thanks{Manuscript received: November, 1, 2024; Revised January, 21, 2024; Accepted February, 20, 2025.}
\thanks{This paper was recommended for publication by
Editor Jens Kober upon evaluation of the Associate Editor and Reviewers’
comments.}
\thanks{$^{1}$The authors are with the Control of Networked Systems Research Group at the Institute of Smart Systems Technologies,
        University of Klagenfurt, 9020 Klagenfurt, Austria
        {\tt\small \{firstname.lastname\}@ieee.org}}

\thanks{Digital Object Identifier (DOI): 10.1109/LRA.2025.3555886}}

\begin{document}

\setlength{\abovedisplayskip}{3pt}
\setlength{\belowdisplayskip}{3pt}
\maketitle
\input{sections/1.Abstract.tex}

\begin{IEEEkeywords}
Parallel Robots, Reinforcement Learning, Model Learning for Control
\end{IEEEkeywords}
\vspace{-0.3cm}

\input{sections/2.Introduction.tex}
\input{sections/3.Related_work.tex}
\input{sections/4.System_description.tex}

\input{sections/5.Simulation_Environments}

\input{sections/6.Controller_Approach}

\input{sections/8.Experiments}

\input{sections/9.Conclusion}

\addtolength{\textheight}{-2cm}

\bibliographystyle{IEEEtran}

\bibliography{bibtex/literatur}

\end{document}

%% file: sections/1.Abstract.tex
\begin{abstract}\label{sec:abstract}
 This paper introduces the Cable Robot Simulation and Control (CaRoSaC) Framework, which integrates a realistic simulation environment with a model-free reinforcement learning control methodology for suspended Cable-Driven Parallel Robots (CDPRs), accounting for the effects of cable sag. Our approach seeks to bridge the knowledge gap of the intricacies of CDPRs due to aspects such as cable sag and precision control necessities, which are missing in existing research and often overlooked in traditional models, by establishing a simulation platform that captures the real-world behaviors of CDPRs, including the impacts of cable sag. The framework offers researchers and developers a tool to further develop estimation and control strategies within the simulation for understanding and predicting the performance nuances, especially in complex operations where cable sag can be significant. Using this simulation framework, we train a model-free control policy rooted in Reinforcement Learning (RL). This approach is chosen for its capability to adaptively learn from the complex dynamics of CDPRs. The policy is trained to discern optimal cable control inputs, ensuring precise end-effector positioning. Unlike traditional feedback-based control methods, our RL control policy focuses on kinematic control and addresses the cable sag issues without being tethered to predefined mathematical models. We also demonstrate that our RL-based controller, coupled with the flexible cable simulation, significantly outperforms the classical kinematics approach, particularly in dynamic conditions and near the boundary regions of the workspace. The combined strength of the described simulation and control approach offers an effective solution in manipulating suspended CDPRs even at workspace boundary conditions where traditional approach fails, as proven from our experiments, ensuring that CDPRs function optimally in various applications while accounting for the often neglected but critical factor of cable sag.
 \looseness=-1
\end{abstract}

%% file: sections/2.Introduction.tex
\section{Introduction}\label{sec:introduction}
CDPRs have emerged as a powerful subset of parallel manipulators, offering enhanced flexibility due to the replacement of rigid links with flexible cables. These robots present a myriad of advantages including expansive workspace, higher payload-to-weight ratios, lower manufacturing costs, portability, and economical construction. As the demand for larger load capacities and workspace in modern engineering grows, CDPRs are becoming increasingly significant. 
CDPRs demonstrate a diverse array of applications, encompassing fields such as logistics\cite{pedemonte2020fastkit}, sports broadcasting\cite{gordievsky2008design}, construction\cite{NISTRobocrane}, rehabilitation\cite{chen2019design}, and extending to specialized domains like agriculture and maritime operations\cite{angelini2023underactuated}.

Despite their growing prominence and diverse applicability, the intricate dynamics of CDPRs, particularly the phenomenon of cable sag, pose significant challenges in achieving precise control and optimal performance. Cable sag, a result of the cables' mass, inherent flexibility, mass of attached payload and gravitational forces, introduces non-linearities and complexities in the system's behavior, complicating accurate modeling and control. Traditional control methodologies often fall short of fully addressing these complexities, leading to compromised precision and efficiency in CDPR operations.

Fundamentally, suspended CDPRs can be fully controlled in terms of position with just three cables if all are attached at the same point, ensuring positional accuracy and stability. However, a fourth cable brings about a significant paradigm shift by introducing redundancy into the system. This redundancy, while seemingly a complication, can significantly enhance the robot's capabilities, offering improved load distribution, enhanced stability, and the potential for fault tolerance.
For precise task-space position control in CDPRs, it's crucial not only to derive the cable lengths from the robot's position, known as the Inverse Kinematics (IK) but also to compute the end-effector's position and orientation from these lengths, termed Forward Kinematics (FK), which is pivotal as it offers real-time feedback on the robot's current state, enabling control algorithms to make necessary adjustments. However, the presence of flexible cables introduces significant non-linearities, complicating the formulation of both FK and IK. This is primarily due to the need to incorporate cable sag effects into these calculations. In practice, this often requires sophisticated models or simulation-based approaches to predict cable behavior under varying loads and positions. 

\begin{figure}[!t]
    \centering
    \fcolorbox{white}{white}{ 
        \includegraphics[width=0.99\columnwidth]{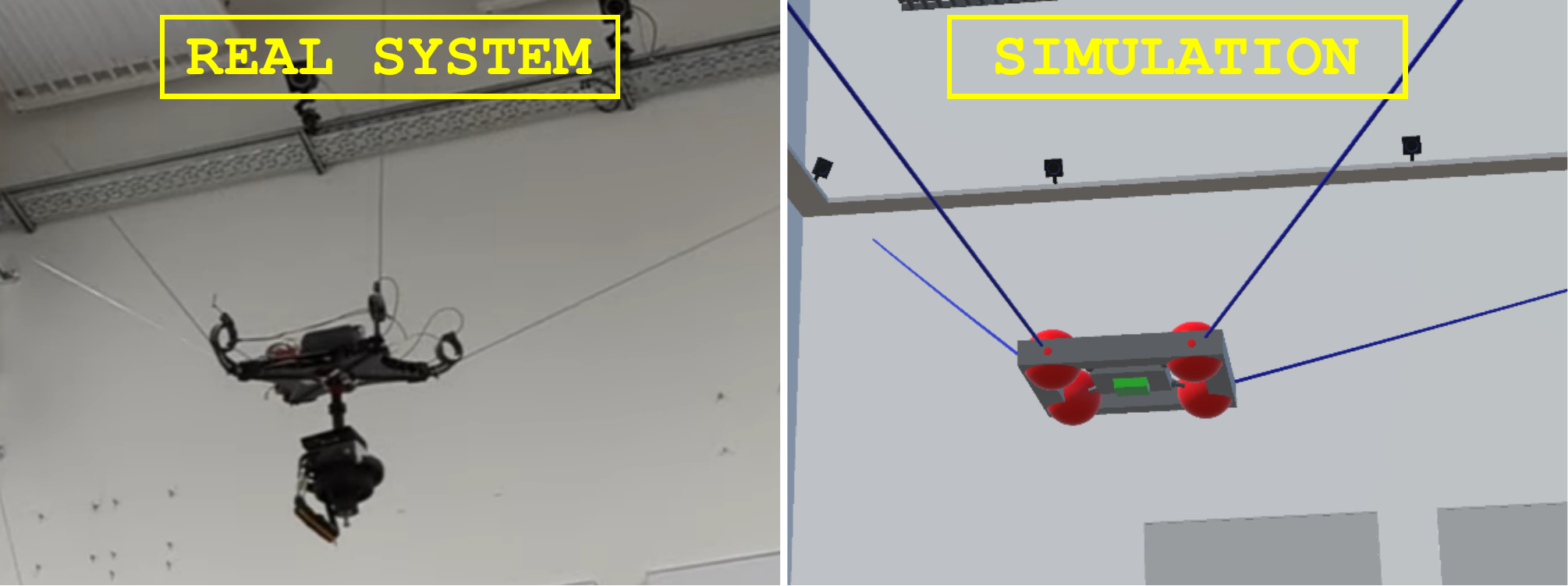}
    }
    \caption{Reference CDPR system (\texttt{Left}) used for setting up our Unity3D simulation (\texttt{Right}) and for generating real-world trajectories.}
    \label{fig:real_sys}
\end{figure}
\vspace{-0.3cm}
    
\subsection{Motivation}
Our primary motivation for adopting a learning-based controller for CDPRs lies in addressing the inherent complexities introduced by sagging cables. Cable sag, caused by the flexibility and mass of cables, leads to non-linear dynamics that classical control methods struggle to handle without extensive modeling, detailed system information, and assumptions that may not hold in varying conditions. Moreover, traditional methods often require remodeling for changes in system configuration, limiting their adaptability in real-world scenarios.
In contrast, a learning-based approach inherently captures these dynamics directly from data, eliminating the need for explicit sag modeling or assumptions about disturbances. By iteratively interacting with the environment, the RL-based controller autonomously learns optimal strategies to manage cable sag and other complexities, even under dynamic and boundary conditions. Unlike methods that compensate for uncertainties, this approach does not rely on highly accurate feedback or predefined models, making it more robust and generalizable. This adaptability ensures precise and reliable control, offering a scalable and practical solution for suspended CDPRs in real-world scenarios.
\looseness=-1
\vspace{-0.4cm}
\newcommand{\SubItem}[1]{
    {\setlength\itemindent{15pt} \item[{\scalebox{0.6}{$\blacksquare$}}] #1}
}

\subsection{Contribution: CaRoSaC Framework}
    \begin{itemize}
        \vspace{0.1cm}
    
        \item {Cable Robot Simulation (CaRoSim) Setup: To the best of our knowledge, we are introducing the first work of a novel simulation setup specifically designed to emulate the real-world behavior of CDPRs with flexible cables exhibiting realistic cable sag behavior.}
        \vspace{0.1cm}

        \item {Learning-based Control Methodology with cable lengths as control inputs}: A model-free control policy based on RL is developed and presented. Unlike traditional control methods, this policy requires minimal system information and is capable of adaptively learning from the CDPR's complex dynamics, leading to precise end-effector positioning without the need for any complex system modeling or additional supplementary system information or external sensors.
        \vspace{0.1cm}

        \item Evaluations on Real-world Data: A comparative assessment of trajectory tracking performance between our learned controller and a classical approach is conducted, demonstrating precision in control.
        Our comprehensive validation of the simulation and learning-based controller using real-world data from an industrial CDPR system showcases promising results. 
        \vspace{0.1cm}
        
        \item Codebase and User Guide: The complete simulation code and a user guide are available as a GitHub repository\footnote{\url{https://github.com/aau-cns/CaRoSim}} to facilitate reproducibility and further research. 
    \end{itemize}
    \vspace{-0.2cm}

%% file: sections/3.Related_Work.tex
\section{Related Work}\label{sec:related_work}
CDPRs have been extensively researched, particularly in achieving precise control for various applications. Traditional control methods for CDPRs often rely on kinematic and dynamic models to determine the required cable lengths and tensions to achieve a specific end-effector position and orientation. However, accurately deriving the kinematics and dynamics of a CDPR system necessitates accounting for the effects of cable sag caused by the cable's self-weight and the nonlinear stress-strain relationships inherent in flexible cables. These factors are crucial for maintaining the precision and reliability of the system in real-world applications.
\vspace{-0.3cm}

\subsection{Classical Control}

Modeling and controlling CDPRs with flexible cables is challenging due to the inherent complexities of the cables, such as mass, elasticity, and sagging under their weight. Accurate control of these systems requires detailed kinematic and dynamic models that account for these factors. Various modeling techniques have been developed to address these complexities. For instance, \cite{7759638} derives the direct kinematics of CDPR using a numerical continuation scheme, based on the Irvine sagging-cable model \cite{irvine1992cable}. This approach enables accurate cable length estimation under sagging conditions.
Additionally, multi-body models based on Cosserat rod theory, as shown in \cite{tempel2019dynamics}, account for large elastic deformations, such as strain and bending, critical for precise control in highly dynamic environments. The work of \cite{1638335} addresses sag-induced cable flexibility by solving the IK problem assuming that the manipulator is either static or moving slowly enough to be considered quasi-static. This method provides a foundation for controlling cable-driven robots with more predictable behavior. The use of port-Hamiltonian methods \cite{schenk2018port} has further enhanced the control by deriving dynamic equations through the total kinetic and potential energies of the system, allowing for energy-based control strategies. In \cite{du2012dynamic} and \cite{miermeister2010modelling}, a linear spring model with mass is used to represent the flexible cables, which allows for the derivation of dynamic equations needed for control. Meanwhile, non-linear behaviors under load, such as cable sag, have been modeled using catenary equations, as described in \cite{erenallak}, offering more realistic representations of the system dynamics under varying load conditions.

Controlling these systems, however, involves more than just accurate modeling. Early control methods \cite{7353995} attempted to solve the nonlinear equations governing the kinematics and dynamics of CDPRs, but they were hampered by computational complexity and system nonlinearity. To overcome these challenges, more sophisticated control techniques have been introduced. For instance, force and torque balance equations have been used to improve control accuracy \cite{GAO201456}, though these approaches come with higher computational demands and the need for precise knowledge of cable sagging behavior.

Model Predictive Control (MPC) has emerged as a powerful tool for controlling CDPRs. For example, the authors in \cite{bettega2023model} developed an MPC-based scheme that uses a nonlinear dynamic model of the CDPR for trajectory planning, followed by inverse dynamics to calculate the required motor torques. This approach allows for anticipatory control actions that consider future states of the system, improving both performance and precision, though it requires extensive dynamic modeling and computational resources. Cascaded control systems have also been explored to enhance control precision. In \cite{8521707}, a multi-loop control system significantly improves end-effector positioning accuracy by integrating visual feedback to adjust for real-time inaccuracies.
\looseness=-1

Accurate kinematic and dynamic models are essential for effective CDPR control, providing a foundation for advanced strategies that address the complexities of flexible cables, ensuring stability and precision. However, these methods often require detailed system knowledge, including cable sag, elasticity, and force distribution, as well as accurate environmental measurements, leading to high computational complexity and implementation effort.
\vspace{-0.3cm}

\subsection{RL Control}

With recent advancements in learning-based approaches, RL has emerged as a promising approach for CDPR control, capable of learning optimal control policies directly from interaction with the environment. This method is particularly appealing for dynamic tasks or when the system model is complex or partially unknown. Studies such as \cite{8948792}, \cite{sancak2022position}, \cite{bouaouda2023dynamic}, and \cite{10035491} demonstrate the potential of RL in controlling CDPRs, showing that it can adapt to various tasks and environments. However, most RL applications for CDPRs overlook challenges like cable sag, which is critical in larger systems. Our research bridges this gap while addressing the limitations of classical methods that depend on extensive external data, complex modeling, and high computational demands. By addressing these challenges, we aim to advance CDPR control with greater precision, adaptability, and efficiency.
\vspace{-0.3cm}

%% file: sections/4.System_Description.tex
\section{SYSTEM DESCRIPTION}\label{sec:sys_description}
Our research employs an industrial suspended CDPR system in 3-D space, which is actuated by four flexible cables suspended from the ceiling. These cables are actuated by winches anchored to the floor, extending upwards to navigate over pulleys mounted on the walls, before finally connecting to the robot. The robot itself is a square platform, with attachment points at each corner for cable connection. The distances between the winches and their corresponding pulleys are predefined and constant. The lengths of the cables are measured between the robot's attachment points to the pulleys' exit points. To accurately determine the exit points at which each cable passes over the pulley, we used a Leica Tachymeter. The robot's position is calculated based on the cable lengths within our industrial CDPR system. Our system is equipped with a low-level PID controller at each motor, which regulates motor inputs to achieve the desired cable lengths as commanded by the RL-based kinematic controller.

For safety reasons the workspace of the robot is constrained to a cubic area with a side length of 4 m. Within this defined workspace, we generated real-world trajectories to test our simulations and evaluate the controller's performance. We used the same workspace within our Unity3D simulation to train the RL agent. Fig.~\ref{fig:real_sys} shows our installed CDPR system. The total weight of the robot is 23.655 kg. We use aramid fiber cables with a mass of 10.55\,g/m and a Young's modulus of $(130\text{–}179)\times 10^9\,\text{N/m}^2$.
\looseness=-1
\vspace{-0.3cm}

%% file: sections/5.Simulation_Environments.tex

\section{SIMULATION ENVIRONMENTS}\label{sec:simulation}
\begin{figure}[!hbt]
    \vspace{-0.3cm}
    \fcolorbox{white}{white}{
            \includegraphics[width=0.9\columnwidth]{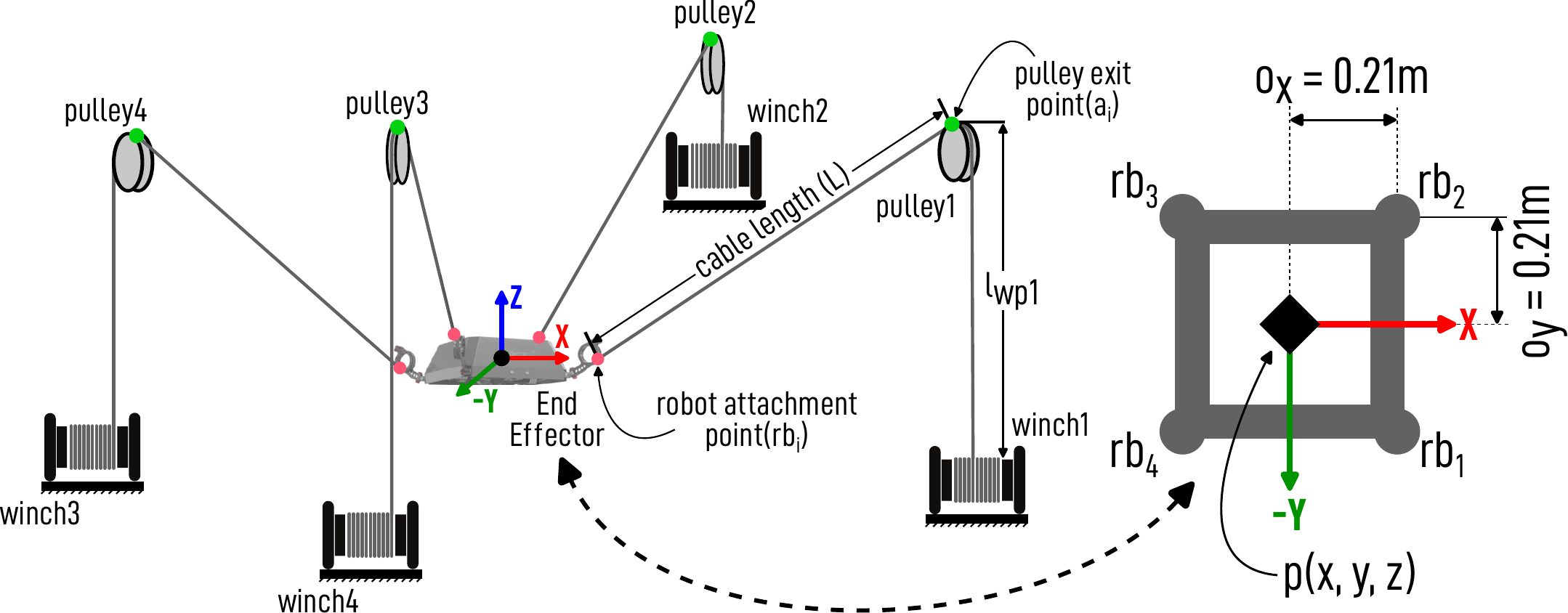}
            }
            \centering
            \caption{Kinematic diagram of our system, depicting all the parameters involved in formulating the kinematics and controller of suspended CDPR.}
            \label{fig:CDPRKine}
    \end{figure}

    We consider two distinct simulation environments in our workflow: the \textit{No-Sag Environment} and the \textit{Cable Robot Simulation (CaRoSim)}. These environments serve complementary purposes, enabling both rapid iteration during development and realistic simulation for training and evaluation. The schematic in Fig  \ref{fig:CDPRKine} shows the kinematics and necessary parameter's information of our system.
    \vspace{-0.4cm}
    \subsection{No-Sag Environment}
    This environment simplifies the system by modeling cables as straight lines, excluding the effects of sag and cable dynamics. The motivation to use a no-sag environment is due to the drastic speed-up in comparison to Unity3D, allowing more rapid iteration over various reward structures and algorithm details without the temporal overhead introduced by the real-time execution of control actions in Unity3D. To obtain the observations that will be needed for training the controller for our no-sag environment, we first formulate the kinematics of the system. 
    \looseness=-1
    \subsubsection{Inverse Kinematics}
     Since we are modeling cables as straight lines, for a given robot position we get the closed-form solution for the cable lengths ($L_i$) as IK by calculating the Euclidean distance between $rb_i$ and ${a}_i$, where $o_i = (o_x, o_y)$ are the offsets from robot's center to the points at which the cables are attached. We use this solution as a ground-truth when evaluating the performance of our RL controller.
    \begin{equation}\label{eq:1}
        L_i = \| (\mathbf{p} + \mathbf{o}_i)  - \mathbf{a}_i \|, \quad i = 1, \ldots, 4 
    \end{equation}
    \subsubsection{Forward Kinematics}
    We formulate the FK problem as an optimization task. Specifically, we aim to minimize the difference between the measured cable lengths (\( L_i \)) and the cable lengths calculated from the initial guessed position. 
    To determine the position \(\mathbf{p}^*\) that minimizes this discrepancy, we opt for Sequential Least Squares Programming to minimize the cable error and solve the following optimization problem:
    \begin{equation}\label{eq:3}
        \mathbf{p}^* = \arg \min_{\mathbf{p}} E(\mathbf{p})
    \end{equation}
    \vspace{-0.5cm}
    \begin{equation}\label{eq:2}
        E(\mathbf{p}) = \sum_{i=1}^{n} \left( \| (\mathbf{p} + \mathbf{o}_i) - \mathbf{a}_i \| - L_i \right)^2
        \vspace{-0.5cm}
    \end{equation}
    \subsection{CaRoSim}
    Our simulation framework uses Unity3D as the base platform because of Unity3D's superior rendering capabilities and extensive support for dynamic soft-body simulations over a dedicated robotic simulation platform like CoppeliaSim or Gazebo. The FK in this environment is determined by the simulation engine, which dynamically resolves the physical interactions between the cables and the robot.
    \begin{figure}[!hbt]
    \vspace{-0.2cm}
            \includegraphics[width=0.7\columnwidth]{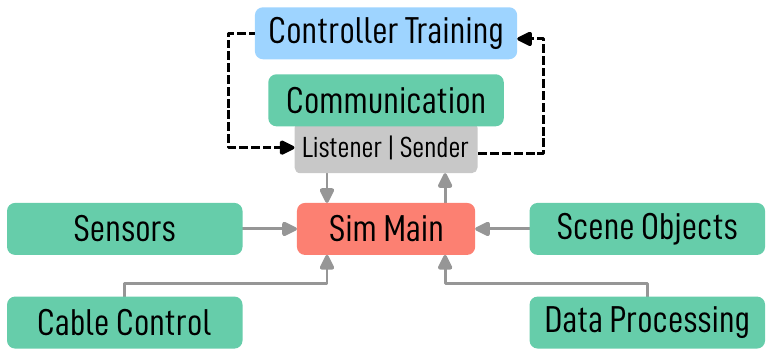}
            \centering
            \caption{Simulation architecture overview, showing different modules and their interactions. The core simulation script manages the individual sub-modules and provides the necessary information for the RL-agent controller.}
            \label{fig:SimFramework}
    \vspace{-0.2cm}
    \end{figure}
    
    To incorporate the cable sag effect, which is paramount in accurately representing the physics of CDPR, we utilized the Obi Rope extension from  Virtual Method Studio\footnote{\url{https://obi.virtualmethodstudio.com/manual/6.1/ropesetup.html}} for Unity3D which offers an accurate, fast, and stable simulation of ropes under various conditions. Its underlying solver is based on Position Based Dynamics \cite{macklin2016xpbd}, a state-of-the-art method for simulating soft bodies, which has been proven to efficiently handle highly dynamic scenarios. Using Obi Rope, we were able to setup our CDPR simulation with intricate interactions between the flexible cables and their environment. Fig. \ref{fig:real_sys} shows our CaRoSim environment scene. We replicated the setup of the real system, ensuring that the positions of pulleys, robot attachment points, and cable exit points in the simulation matched their physical counterparts. Additionally we modeled the total weight of the CDPR, including the attached payload, as a single entity and shifted the center of gravity to account for this unified weight distribution. This adjustment helps simplify the simulation by reducing complexity.

    Fig.~\ref{fig:SimFramework} shows the architecture of our CaRoSim environment consisting of several modules to facilitate a plug-and-play setup that promotes ease of use and extensibility. Each module is designed to operate both independently and as part of a cohesive whole, providing a flexible environment for testing various estimation and control strategies and robot dynamics.
    \vspace{-0.5cm}

%% file: sections/6.Controller_Approach.tex
\vspace{-0.2cm}
\section{Controller Approach}\label{sec:methods}
     To develop a robust controller for the CDPR that accounts for cable sag, we adopt a staged RL training process. Initially, we train in a simplified \textit{no-sag environment}, which allows rapid experimentation with various reward structures and algorithm parameters due to its computational efficiency. Once the RL controller shows promising performance under this simplified setup, we advance to the \textit{CaRoSim} environment, where physical effects such as cable sag are accurately simulated using Unity3D. This two-stage training approach enables us to progressively refine the controller, validating foundational elements like reward structure and control stability in the no-sag model before addressing the added complexity of real-world physics in CaRoSim. By training in a less complex environment first, we mitigate the computational demands of high-fidelity simulations, streamlining the initial development phase.
    \looseness=-1
    \begin{table}[!hbt]
    \small
    \vspace{-0.3cm}
    \centering
    \caption{Environment State and Actions}
        \begin{tabular}{|c|c|c|}
        \hline
        Parameters & Contents & Dimension \\ 
        \hline
        \hline
        \multirow{3}{*}{State}      & \multirow{1}{*}{Current cable lengths} 
                                    & \multirow{1}{*}{4x1}  \\ \cline{2-3}
                                    
                                    & \multirow{1}{*}{Current robot position} 
                                    & \multirow{1}{*}{3x1}  \\ \cline{2-3}
                                    
                                    & \multirow{1}{*}{Target robot position}  
                                    & \multirow{1}{*}{3x1}  \\ \cline{2-3}
        \hline
        \hline
        \multirow{1}{*}{Actions}   & \multirow{1}{*}{Cable lengths: $[L_1, L_2, L_3, L_4]$} & \multirow{1}{*}{4x1}  \\
        \hline
        \end{tabular}
    \label{tabel: Static State}
    \end{table}

    Table ~\ref{tabel: Static State} shows the state and action spaces used to train a learning-based controller for CDPR. Our state consists of cable lengths, robot position, and target position, where cable lengths predicted by the learning agent act as the control inputs, robot position is the environment observation, and the target position is provided as a task objective.

    The RL-based controller uses cable lengths as its action space, abstracting low-level motor control to focus on the kinematic relationship between cable lengths and end-effector position. This simplifies control while addressing the complexities of flexible cables and sagging. The industrial CDPR system executes the cable length commands using low-level PID controllers to regulate motor inputs, ensuring precise end-effector positioning, as detailed in Section \ref{sec:sys_description}.
      
\begin{figure}[!hbt]
    \vspace{-0.2cm}
    \centering
      \includegraphics[width=0.85\columnwidth]{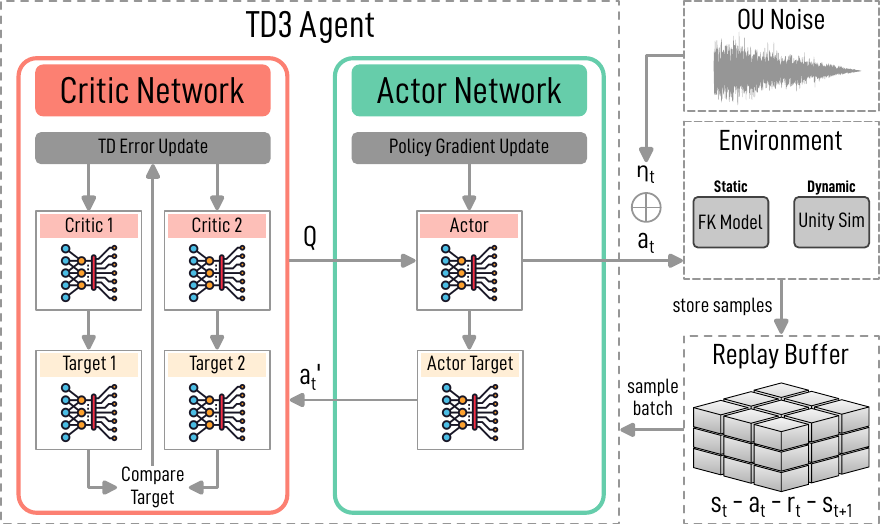}
    \caption{Overview of the architecture used for training a TD3 agent in a no-sag and CaRoSim environment.}
    \label{fig:TD3}
    \vspace{-0.5cm}
\end{figure}

    As RL controller we used Twin Delayed Deep Deterministic Policy Gradient algorithm (TD3)\cite{fujimoto2018addressing}, Fig.~\ref{fig:TD3} shows the architecture of the algorithm used to train our controller. We trained a TD3 agent to learn a task-space position tracking controller. The informed choice of the RL algorithm was made due to its suitability for continuous control environments. TD3's architecture with twin Q-networks and delayed policy updates, significantly enhances the learning stability by effectively mitigating the overestimation of bias commonly observed in its predecessor Deep Deterministic Policy Gradient (DDPG) algorithm \cite{lillicrap2016continuous}. TD3's off-policy learning paradigm capitalizes on each data sample more effectively than on-policy counterparts like Proximal Policy Optimization (PPO)\cite{schulman2017proximal}, thus expediting the training process without sacrificing performance. Additionally, TD3's target policy smoothing technique adeptly addresses the noise and uncertainties characteristic of CDPR systems, ensuring the learned controller's resilience to minor disturbances or modeling inaccuracies.

    \begin{align}
        r^{\mathrm{full}}_{\mathrm{t}} &= r^{\mathrm{step}}_{\mathrm{t}} + r^{\mathrm{dist}}_{\mathrm{t}} + r^{\mathrm{goal}}_{\mathrm{t}} \\
        r^{\mathrm{step}}_{\mathrm{t}} &= +0.01\\
        r^{\mathrm{dist}}_{\mathrm{t}} &= -\tanh(2 * (d_{\mathrm{norm}} - 0.06)) + 0.002\\
        r^{\mathrm{goal}}_{\mathrm{t}} &= 
            \begin{cases}
                +1.0,  & \text{if, $g_{dist} \leq  0.05m$}\\
                0, & \text{if, $g_{dist} >  0.05m$}\\
            \end{cases}
    \end{align}

    We define our reward function as a cumulative sum of a constant reward $r^{step}_{t}$, a distance reward $r^{dist}_{t}$ and a success reward $r^{goal}_{t}$ for each time-step. The $r^{step}_{t}$ encourages the agent to explore more of the state space and to balance the reward signal, whereas $r^{dist}_{t}$ navigates the agent to reach the target state. For $r^{dist}_{t}$ we use normalized value ($d_{norm}$) of the distance to target position, which is the ratio of the Euclidean distance between current and target positions ($g_{dist}$) to the maximum possible distance within the training workspace. We assign a high positive reward of $r^{goal}_{t}$ when the robot position is within the distance threshold of 0.05m. The reward structure with the selected constants is formulated to have a generic behavior for a target-reaching task and thus can be easily transferred to different setup configurations with simple scale adjustments.
    \looseness=-1
    %

    %
    Our TD3 agent is trained for a total of 5000 episodes in the no-sag environment. At the start of each episode, we set a random cable configuration and define a random target position. We apply the action for each step and get the robot position observation using the FK model. For an initial 100 episodes, we randomly sample the actions from the action space and acquire corresponding observations to populate the replay buffer with diverse training experiences. We start the policy training for the remaining number of episodes, where the continually learning policy predicts an action for a given state. To facilitate exploration during the initial phase of training, we apply a decaying Ornstein-Uhlenbeck process noise (OUNoise) to the actions chosen by the agent based on its current policy state. This decay mechanism is tailored to the progression of training episodes, ensuring that the agent engages more in exploration early on, gradually shifting towards exploiting the knowledge it has acquired in later episodes. The decay rate is designed to balance the need for new information in the early stages with the benefit of leveraging learned behaviors as the agent becomes more experienced. The algorithm parameters for the training are presented in Table~\ref{table: hyperparameters}.
    \looseness=-1
    
\begin{table}[!hbt]
    \small
    \vspace{-0.4cm}
    \centering
    \caption{TD3 Hyper-parameters}
        \begin{tabular}{|c|c|l|}
        \hline
            Parameters                  & Variable              &Values \\
            \hline
            \hline
            Number of episodes          &$n_{episodes}$          &$5000$  \\
            \hline
            Number of steps             &$n_{steps}$            &$200$  \\
            \hline
            Batch size                  &$n_{batch}$              &$256$  \\
            \hline
            Actor learning rate         &$lr_{ac}$                &$3e-04$  \\
            \hline
            Critic learning rate        &$lr_{cr} $               &$3e-03$  \\
            \hline
        \end{tabular}
    \label{table: hyperparameters}
    \end{table}

    After validating the working principle of our TD3 components in the no-sag environment, we moved to training the TD3 agent within the dynamic CaRoSim environment to encapsulate the full spectrum of system dynamics - including robot and cable mass, cable sag, gravitational effects, and actuation forces. This shift was instrumental in preparing the controller learning for real-world application, allowing it to adapt to and learn from the complex behaviors inherent in CDPR systems.
    \looseness=-1
    
    The state-space and action-space for the CaRoSim environment in Unity3D are similar to that of the no-sag environment, as shown in Table~\ref{tabel: Static State}. However, the complexity of the reward structure increases to capture the dynamics and learn a policy that efficiently performs task-space control in the presence of all the dynamical effects.
    \looseness=-1
    
    In addition to the rewards used in no-sag environment $r^{step}_{t}$, $r^{dist}_{t}$, $r^{goal}_{t}$, the cumulative reward function for the CaRoSim environment consists of, sag reward $r^{csag}_{t}$, change reward $r^{cact}_{t}$ and deviation reward $r^{cdev}_{t}$. The sag reward ($r^{csag}_{t}$) penalizes actions causing excessive cable sag. 
    \vspace{0.2cm}
    %
    \begin{align}
        r^{\mathrm{full}}_{\mathrm{t}} &= r^{\mathrm{step}}_{\mathrm{t}} + r^{\mathrm{dist}}_{\mathrm{t}} 
        + r^{\mathrm{csag}}_{\mathrm{t}} + r^{\mathrm{cact}}_{\mathrm{t}} + r^{\mathrm{cdev}}_{\mathrm{t}} \\
        r^{\mathrm{step}}_{\mathrm{t}} &= +0.01\\
        r^{\mathrm{dist}}_{\mathrm{t}} &= -\tanh(2 \times (d_{\mathrm{norm}} - 0.06)) + 0.002\\
        r^{\mathrm{csag}}_{\mathrm{t}} &= -\tanh(c_{diff}^2 + 0.09) \times 0.7 - 0.035\\
        r^{\mathrm{cact}}_{\mathrm{t}} &= -\tanh(c_{ratio})^{15} / 10\\
        r^{\mathrm{cdev}}_{\mathrm{t}} &= -\tanh(a_{diff})^{4} / 5\\
        r^{\mathrm{goal}}_{\mathrm{t}} &= 
            \begin{cases}
                +1.0,  & \text{if, $g_{dist} \leq  0.05m$}\\
                0, & \text{if, $g_{dist} >  0.05m$}\\
            \end{cases}
    \end{align}
    
\begin{figure}[!hbt]
    \vspace{-0.2cm}
        \includegraphics[width=0.9\columnwidth]{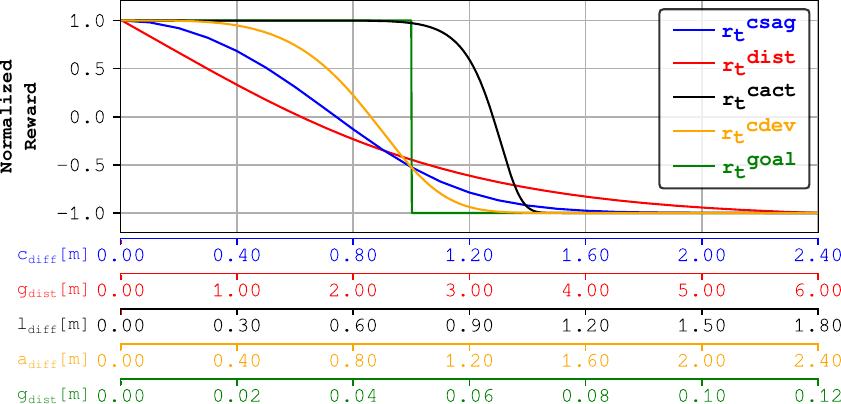}
        \centering
        \caption{The behavior of all reward components described in Section \ref{sec:methods} is depicted. Each reward function is plotted against its respective dependent variable, shown on multiple x-axes with corresponding colors matching the reward curves. This visualization demonstrates how each reward depends on specific global information about the system.}
        \label{fig:reward_behavior}
    \vspace{-0.2cm}
    \end{figure}

    We parameterize cable sag by current cable lengths and respective straight-line distances between the cable exit point and robot attachment point for each cable where we use the difference between the actual cable length and straight-line distance ($c_{diff}$) to quantify the cable sag. The change reward ($r^{cact}_{t}$) is designed to make the agent predict actions based on the ratio ($c_{ratio}$) of the difference between predicted action and current cable length ($l_{diff}$) to the distance to goal ($g_{dist}$), discouraging disproportionately large or small actions. Lastly, the deviation reward ($r^{cdev}_{t}$) encourages the agent to minimize cable lengths, promoting a straight-line approach to cable configuration while allowing flexibility for boundary conditions. We formulate $r^{cdev}_{t}$ based on absolute difference ($a_{diff}$) between the possible straight-line cable lengths using IK ($c^{ik}_{t}$) and the predicted action from our TD3 agent ($a_t$) for a given target robot position. Each reward component plays a distinct role in shaping the agent's decision-making process, balancing efficiency with the physical constraints of cable configuration. Similar to the reward structure of the no-sag environment, we designed the rewards for CDPR with flexible cables to have generic behavior based on global information of the setup making it portable to other setup configurations, Fig. \ref{fig:reward_behavior} shows the behavior of all individual reward functions.

    The TD3 hyperparameters for training in Unity3D differ from the no-sag environment only in the number of episodes and steps per episode. The agent was trained in Unity3D for 1000 episodes with 30 steps each (30,000 steps), significantly lower than the no-sag setup of 5000 episodes with 200 steps each, due to the time required for actuation and the deterioration of floating-point precision in Unity3D over longer durations, which affected cable actuation behavior reliant on simulation frame intervals. Unity3D training took 16 hours for 30,000 steps, compared to 2 hours for the same steps in the no-sag environment. Training was conducted on an Intel Core i7-11700KF processor (3.60 GHz, 16 threads) and an NVIDIA GeForce RTX 3060 GPU with 12GB VRAM.
\vspace{-0.2cm}

%% file: sections/8.Experiments.tex
\section{RESULTS}\label{sec:experiments}
This section presents the results in three parts: first, we validate the accuracy of the simulation framework by comparing it with real-world data. Next, we analyze the training performance of the RL controller in both the No-Sag and CaRoSim environments. Finally, we evaluate the controller's performance across the No-Sag, CaRoSim, and real-world environments to demonstrate its adaptability to sagging and non-linear dynamics. In real experiments, we obtain the robot position and cable length information using a motion capture system and the internal sensors of our CDPR setup, respectively. In simulations, this information is derived from the scene object properties within the environment. For all experiment results, Ground Truth (GT) represents the reference robot position. Specifically, GT corresponds to the motion capture data in real-world experiments and the simulated positions in CaRoSim. The trajectories used in our experiments ensure smooth motion. Real-world experiments employed minimum snap trajectories with speeds from \textbf{0.0 m/s to 2.08 m/s}, while simulations used cubic spline trajectories with speeds ranging from \textbf{0.0 m/s to 5.0 m/s} for controller evaluations.
\looseness=-1

\vspace{-0.1cm}

\subsection{Simulation Validation}
\begin{figure}[!hbt]
    \vspace{-0.2cm}
            \includegraphics[width=0.9\columnwidth]{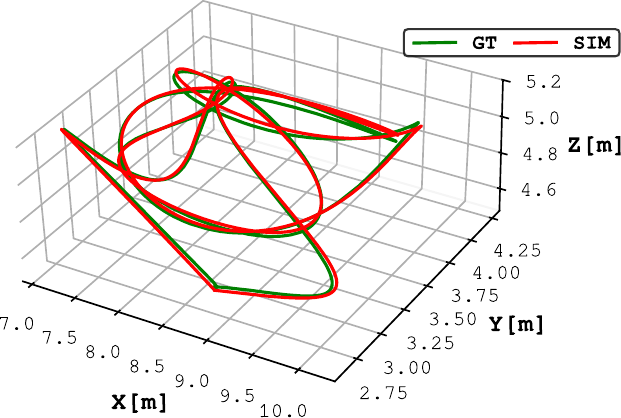}
            \centering
            \caption{Comparison of a real CDPR system's 3D trajectory (\texttt{GT}) and our simulation framework's output (\texttt{SIM}) based on recorded cable lengths.
            \looseness=-1
            }
            \label{fig:SimEval}
            \vspace{-0.2cm}
    \end{figure}

    To validate the accuracy and functionality of our simulation setup, we conducted evaluations using real-world data. This involved setting cable lengths from real-world data, into CaRoSim and subsequently recording the simulated robot's position and applied cable lengths. These positions and cable lengths from the simulation were then compared to the corresponding real-world positions of the robot and cable lengths. Fig.~\ref{fig:SimEval} shows the close alignment of our simulation trajectory, marked in red, with the real-world trajectory, in green. 
    
    \begin{figure}[!hbt]
            \includegraphics[width=1\columnwidth]{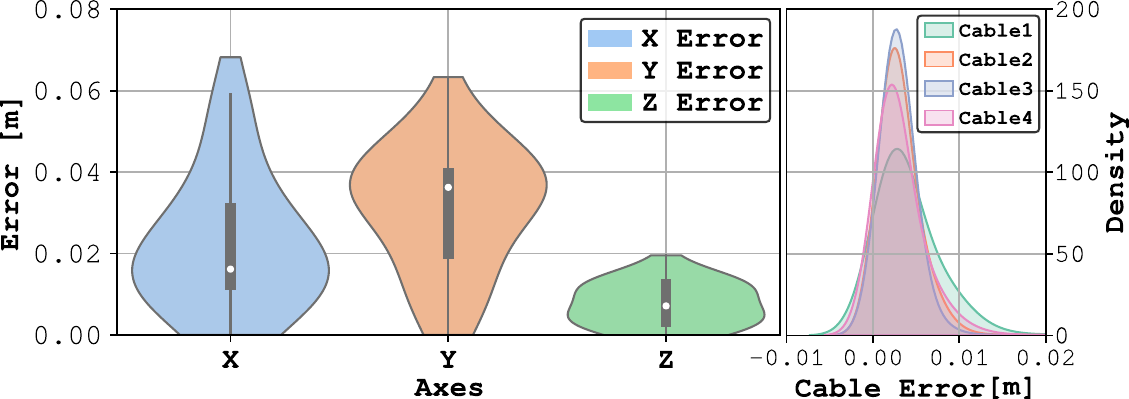}
            \centering
            \caption{Violin plot (left) showing the distribution of absolute errors in position for the X, Y, and Z axes when comparing real-world trajectories with those output by our simulation framework. Cable error distribution plot (right) displays the error distribution in setting cable lengths.}
            \label{fig:SimEval_error}
         \vspace{-0.225cm}
    \end{figure}
    
    Whereas Fig.~\ref{fig:SimEval_error} provides a detailed analysis by presenting the position error distribution along the X, Y, and Z axes, highlighting the comparison between the real-world trajectory and the trajectory generated by our simulation framework. The Root Mean Squared Error (RMSE) of the CaRoSim output on the evaluated trajectory is \textbf{0.027m} for the X-axis, \textbf{0.034m} for the Y-axis, and \textbf{0.009m} for the Z-axis. 
    Whereas the mean absolute errors are \textbf{0.0226m}, \textbf{0.0312m}, and \textbf{0.0076m} for the X, Y, and Z axes, respectively. These results underscore the high precision and fidelity of our simulation framework in replicating real-world movements, with mean absolute error under \textbf{1\%} for X and Z axes and under \textbf{2\%} for Y axis relative to their respective workspace spans. This highlights the precision with which our simulation framework replicates the actual movements of the robot, making it a promising platform for controller development and other experimental simulations. The cable length error in the simulation is computed as the absolute difference between the desired cable length provided as a control input and the actual cable length achieved by the simulation solver.
    \looseness=-1

\subsection{Controller Training}
\begin{figure}[!hbt]
        \includegraphics[width=0.92\columnwidth]{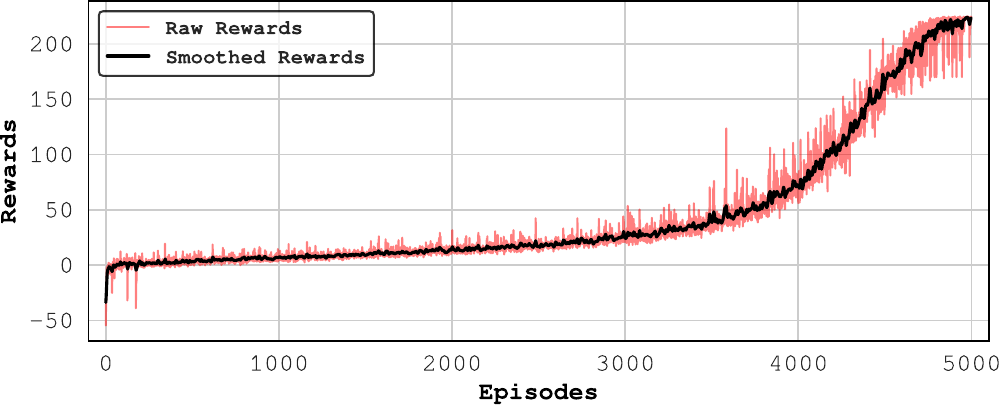}
        \centering
        \caption{Cumulative reward of the TD3 agent for the no-sag environment. Using the smoothed rewards (\texttt{black}) for visualization, we can clearly see the upward trend in the episode rewards.}
        \label{fig:mathrl_rewards}
        \vspace{-0.1cm}
    \end{figure}

    Fig.~\ref{fig:mathrl_rewards} shows the cumulative reward acquired by the agent throughout its training episodes in the no-sag environment.
     
\begin{figure}[!hbt]
        \vspace{-0.2cm}
            \includegraphics[width=0.92\columnwidth]{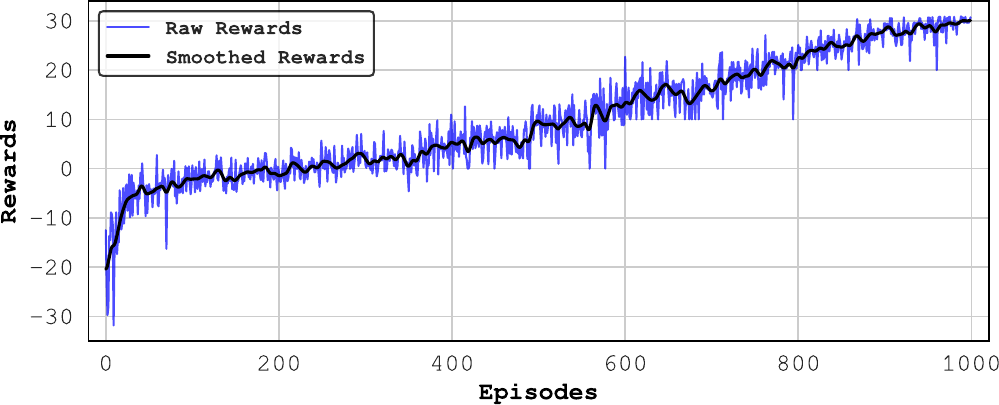}
            \centering
            \caption{Cumulative reward of the RL agent for the CaRoSim environment illustrating the agent's continuous learning and improvement in performance, as evidenced by the gradually increasing rewards over 1000 episodes.}
            \label{fig:simrl_rewards}
        \end{figure}

    Initially, during the first 100 episodes, there is a high degree of variability in the episode rewards. This period corresponds to the random sampling phase, where the agent explores the environment with randomly sampled actions without any learned policy guiding its actions. As the agent transitions into the policy training phase, the episode rewards steadily yield higher returns reaching a plateau towards the end of the training, confirming the convergence of TD3 policy. Fig.~\ref{fig:simrl_rewards} illustrates the TD3 agent's reward trend in the CaRoSim environment, starting with high variance in the initial exploratory phase without a set policy, leading to early reward fluctuations. Over time, a consistent increase in rewards reflects the agent's enhanced decision-making skills. Occasional drops in rewards signal continuous exploration to evade sub-optimal strategies. The curve's eventual leveling off indicates the agent has established a robust policy, marking the model as effectively trained.
    \looseness=-1
    
    \vspace{-0.6cm}

    \subsection{Controller Validation}
        \subsubsection{No Sag Environment}
\begin{figure}[!hbt]
                \includegraphics[width=0.8\columnwidth]{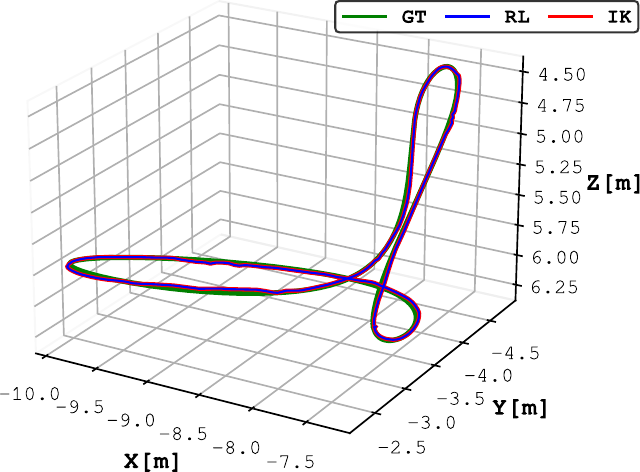}
                \centering
                \caption{3D trajectory comparison between the simulated ground-truth (\texttt{GT}), RL controller (\texttt{RL}) and the classical IK model (\texttt{IK}).}
                \label{fig:MathRL_eval_traj}
                \vspace{-0.1cm}

        \end{figure}

        To validate the learned controller in no-sag environment, we compare the trajectory-tracking performance of our trained TD3 agent against the IK model. We ran a comparison over 1000 random trajectories to establish a statistical evaluation of the comparison. Fig.~\ref{fig:MathRL_eval_traj} depicts a simulated 3D trajectory followed by our TD3 agent, shown in blue (RL), and the IK model shown in red (IK).
        
\begin{figure}[!hbt]
        \vspace{-0.1cm}
                \includegraphics[width=0.95\columnwidth]{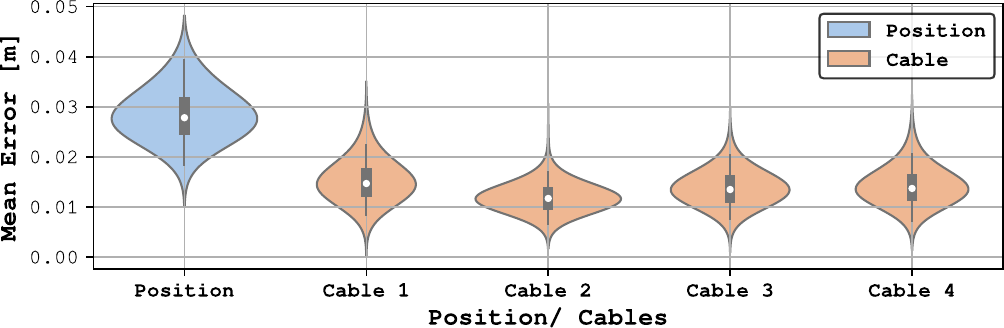}
                \centering
                \caption{Distribution of mean position error (euclidean distance) and mean of absolute cable length errors across 1000 trajectories. The validation demonstrates consistent performance in both position and cable length accuracy with a median of \textbf{0.027m} in position error.}
                \label{fig:MathEval_1000}
        \end{figure}

         The visual representation clearly reveals the ability of the TD3 agent to track the desired trajectory with a degree of precision comparable to that of the classical IK approach. The close alignment of the IK and RL trajectories underscores the TD3 agent's adept position tracking, where the IK model serves as a standard for comparing performance metrics in the absence of cable sag. Fig.~\ref{fig:MathEval_1000} shows the distribution of mean error in robot position and cable lengths for 1000 trajectories. 
        \looseness=-1

        \vspace{-0.2cm}
        \subsubsection{CaRoSim Environment}
\begin{figure}[!hbt]
                \includegraphics[width=0.8\columnwidth]{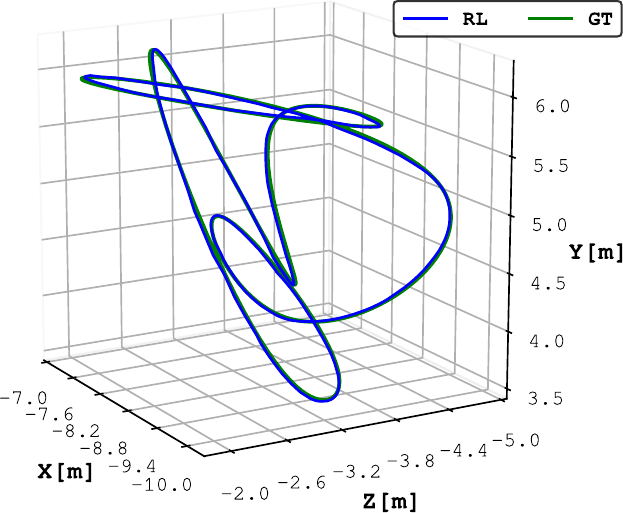}
                \centering
                \caption{3D trajectory comparison in dynamic CaRoSim environment between the RL controller in (\texttt{RL}) and the ground truth trajectory (\texttt{GT}).}
                \label{fig:UnityRLEval_traj}
        \end{figure}

        To further validate the RL controller in a dynamic CaRoSim environment in Unity3D, we conducted trajectory-tracking experiments over 20 trajectories and compared the performance against the GT. The comparison shows how closely the RL controller is able to follow the desired trajectory. Fig.~\ref{fig:UnityRLEval_traj} illustrates a sample trajectory in 3D space, with the RL-tracked trajectory shown in blue and the GT shown in green. The close alignment between the RL and GT trajectories visually demonstrates the controller's high performance in dynamic conditions.
        \looseness=-1
        
        \begin{figure}[!hbt]
                \includegraphics[width=0.95\columnwidth]{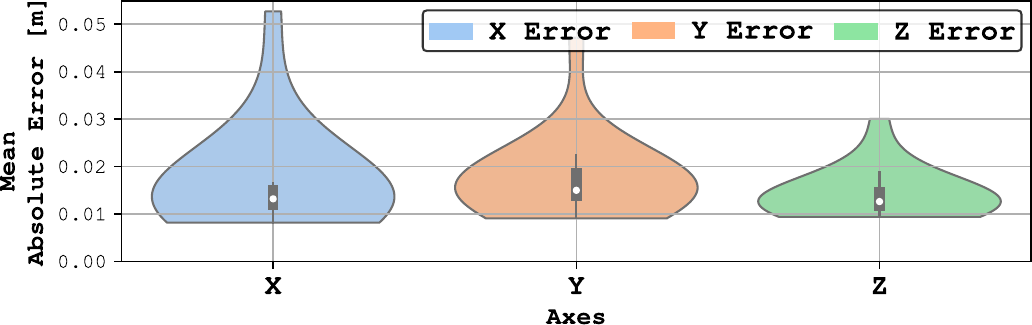}
                \centering
                \caption{Violin plot showing the mean absolute error across 20 trajectories tracked by the RL controller in the CaRoSim environment, with median errors of 0.013m, 0.014m, and 0.012m in the X, Y, and Z axes, respectively.}
                \label{fig:unityRLStats}
        \end{figure}

        Fig.~\ref{fig:unityRLStats} provides a statistical evaluation of errors across 20 trajectories, showing the mean error in each axis and highlighting the RL controller’s precision in trajectory tracking within the dynamic CaRoSim simulation.
        \vspace{0.1cm}
        
        \subsubsection{Real-World Trajectory}
        Finally, we compared our RL controller against the kinematics solver (KS) approach from \cite{erenallak}, which was developed for a similar suspended CDPR system. Their proposed solver incorporates cable sag by modelling flexible cables as catenary curve and fuses it with an IMU through an Extended Kalman Filter (EKF) to enhance pose estimation accuracy during dynamic motions.
        \vspace{0.1cm}
    
\begin{figure}[!hbt]
                    \includegraphics[width=0.85\columnwidth]{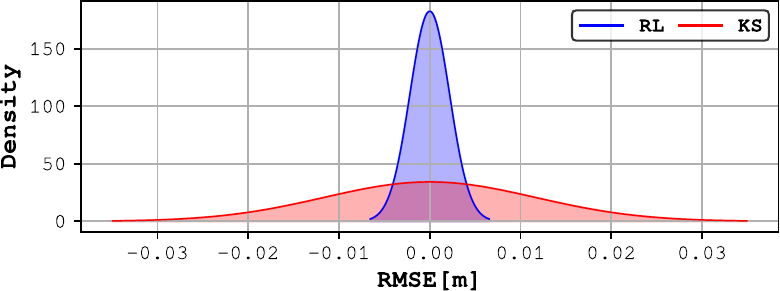}
                    \centering
                    \caption{Gaussian distributions of RMSE for the RL controller (\texttt{RL}) and the kinematic solver approach (\texttt{KS}) across 10 trajectories. The RL controller shows a narrow and consistent error distribution, reflecting lower variability in its RMSE values with a smaller standard deviation of 0.00218m. In contrast, KS with a larger standard deviation of 0.01164m exhibits greater variability in RMSE values.}
                    \label{fig:RealRLEval_dist}
            \end{figure}

        Fig.~\ref{fig:RealRLEval_dist} compares the Root Mean Square Error (RMSE) of both approaches across multiple trajectories, showcasing RL with more consistent performance with errors tightly clustered around the mean. In contrast, the KS exhibits less consistent performance, with a wider spread of error values.
        
            \begin{figure}[!hbt]
                    \includegraphics[width=0.9\columnwidth]{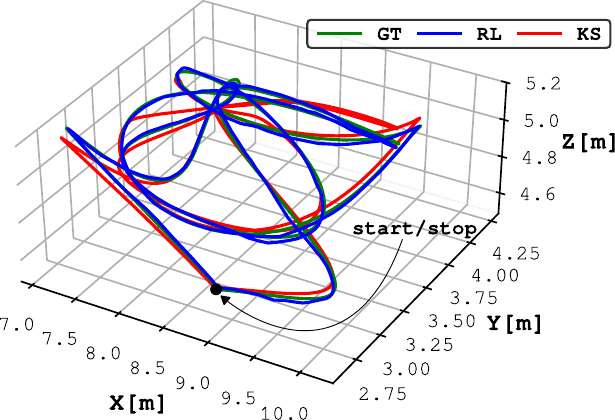}
                    \centering
                    \caption{3D trajectory comparison between the ground truth (\texttt{GT}), RL controller (\texttt{RL}), and kinematic solver (\texttt{KS}). The RL controller demonstrates strong tracking performance with minimal deviation from the ground truth, while the kinematic solver exhibits more significant errors.}
                    \label{fig:RealRLEval3D}
            \end{figure}
    
        In Fig.~\ref{fig:RealRLEval3D} we show the 3D trajectory comparison, where the RL controller (blue) closely follows the ground truth (green) with minimal deviation, whereas the KS approach (red) struggles, particularly at trajectory turns and areas involving dynamic oscillations. The KS approach, while capable of estimating poses, is hindered by the inherent challenges of modeling cable sag and the assumptions made regarding the distribution of vertical forces across cables. These limitations become more prominent at the boundary regions of the operational space, where the performance of the solver degrades.
        
\begin{figure}[!hbt]
                    \vspace{-0.25cm}
                    \includegraphics[width=0.9\columnwidth]{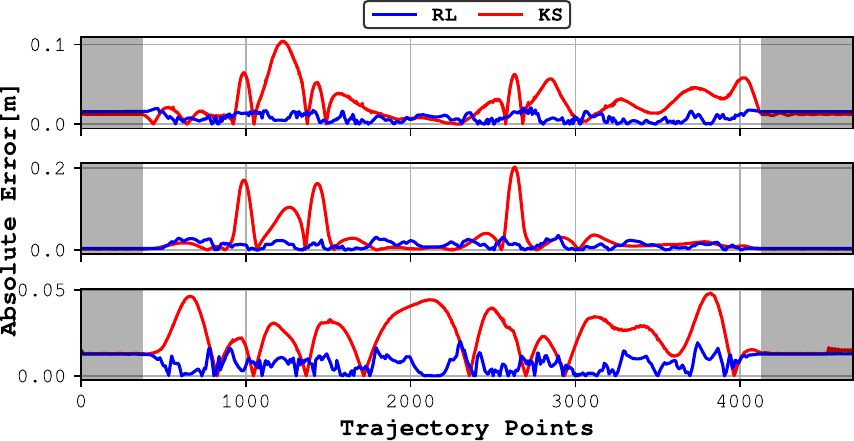}
                    \centering
                    \caption{Absolute tracking error across the X, Y, and Z axes for both the RL controller (\texttt{RL}) and the KS approach (\texttt{KS}). The shaded grey sections are the no-motion segments in the trajectory. Our RL controller maintains consistently low error across all axes, whereas the KS approach exhibits significantly larger oscillations, especially in the Z-axis, reflecting difficulties in handling boundary conditions.}
                    \label{fig:RealRLEval_error}
                    \vspace{-0.6cm}
            \end{figure}

        Fig.~\ref{fig:RealRLEval_error} depicts the absolute tracking error in the X, Y, and Z axes across a trajectory. As seen from the plots, the RL controller maintains low error levels consistently across all axes. In contrast, the KS approach exhibits higher oscillations, particularly in the Z-axis, where errors are most pronounced. This discrepancy highlights the RL controller’s ability to maintain better accuracy during dynamic motions. The RMSE for RL is \textbf{0.023m}, \textbf{0.032m} and \textbf{0.009m}, whereas for KS is \textbf{0.031m}, \textbf{0.044m}, \textbf{0.024m} across X, Y and Z axes respectively.
        \looseness=-1
    
        Overall, the RL controller significantly outperforms the KS approach in terms of trajectory tracking precision, especially towards boundary conditions. The results validate the effectiveness of the RL approach in adapting to complex dynamics without needing assumptions about force distributions or cable behaviors, which are critical in systems with sagging cables.
        \looseness=-1
    
%% file: sections/9.Conclusion.tex
\section{CONCLUSION}\label{sec:conclusion}
This work introduced a novel simulation setup for CDPRs that accurately models cable sag effects and closely replicates real-world dynamics, validated against actual CDPR trajectories. By eliminating the need for conventional matrix formulations and sag computations, our approach simplifies CDPR deployment and operation.

We also proposed a learning-based control strategy for precise target acquisition tasks, effectively capturing the system's intrinsic dynamics. Trajectory-tracking experiments showed that our RL controller consistently outperformed the classical kinematic solver, particularly near workspace boundaries where the solver struggled with force distribution assumptions. The RL controller demonstrated superior accuracy and adaptability, establishing itself as a robust and flexible alternative for advanced CDPR control.
\vspace{-0.4cm}